\documentclass{article}

\usepackage{microtype}
\usepackage{graphicx}
\usepackage{subfigure}
\usepackage{booktabs} 
\usepackage{algorithm}
\usepackage{algorithmic}
\usepackage{amsmath}
\usepackage{tabularx}
\usepackage{makecell}
\usepackage{titlesec}
\usepackage{enumitem}

\usepackage{hyperref}

\usepackage[accepted]{icml2025}

\icmltitlerunning{Return of the Encoder: Maximizing Parameter Efficiency for SLMs}

\begin{document}

\twocolumn[
\icmltitle{Return of the Encoder: Maximizing Parameter Efficiency for SLMs}
\icmlsetsymbol{equal}{*}
\begin{icmlauthorlist}
\icmlauthor{Mohamed Elfeki}{equal,to}
\icmlauthor{Rui Liu}{equal,to}
\icmlauthor{Chad Voegele}{equal,to}
\end{icmlauthorlist}

\icmlaffiliation{to}{Microsoft}

\icmlcorrespondingauthor{Mohamed Elfeki}{melfeki@microsoft.com}

\icmlkeywords{Machine Learning, ICML}

\vskip 0.3in
]

\begin{abstract}
The dominance of large decoder-only language models has overshadowed encoder-decoder architectures, despite their fundamental efficiency advantages in sequence processing. For small language models (SLMs) - those with 1 billion parameters or fewer - our systematic analysis across GPU, CPU, and NPU platforms reveals that encoder-decoder architectures achieve 47\% lower first-token latency and 4.7x higher throughput compared to decoder-only models on edge devices. These gains may be attributed to encoder-decoder's one-time input processing and efficient separation of understanding and generation phases.

We introduce a novel knowledge distillation framework that enables encoder-decoder models to leverage capabilities from large scalable decoder-only teachers while preserving their architectural advantages, achieving up to 6 average performance points improvement across diverse tasks, with significant gains in asymmetric sequence tasks where input and output distributions can benefit from different processing approaches.

When combined with modern advances like Rotary Positional Embeddings (RoPE) and Vision encoders, our systematic investigation demonstrates that encoder-decoder architectures provide a more practical path toward deploying capable language models in resource-constrained environments. Our findings challenge the prevailing trend toward decoder-only scaling, showing that architectural choices become increasingly crucial as parameter budgets decrease, particularly for on-device and edge deployments where computational efficiency is paramount. 

\end{abstract}

\ifdefined\isaccepted
    \par\vspace{-0.5cm}
    \rule{0.8\columnwidth}{0.5pt}
    \vspace{0.2cm}
    \begin{minipage}{\columnwidth}
    \footnotesize
    \textsuperscript{*}Equal contribution. \textsuperscript{1}Microsoft.\\
    Code and models will be released at:\\   \href{https://github.com/microsoft/encoder-decoder-slm/tree/main}{\hspace{0.25cm}microsoft/encoder-decoder-slm}.\\
    \end{minipage}
\else
    \printAffiliationsAndNotice{\icmlEqualContribution}
\fi

\section{Introduction}
\begin{figure*}[t]
    \centering
    \includegraphics[width=\textwidth]{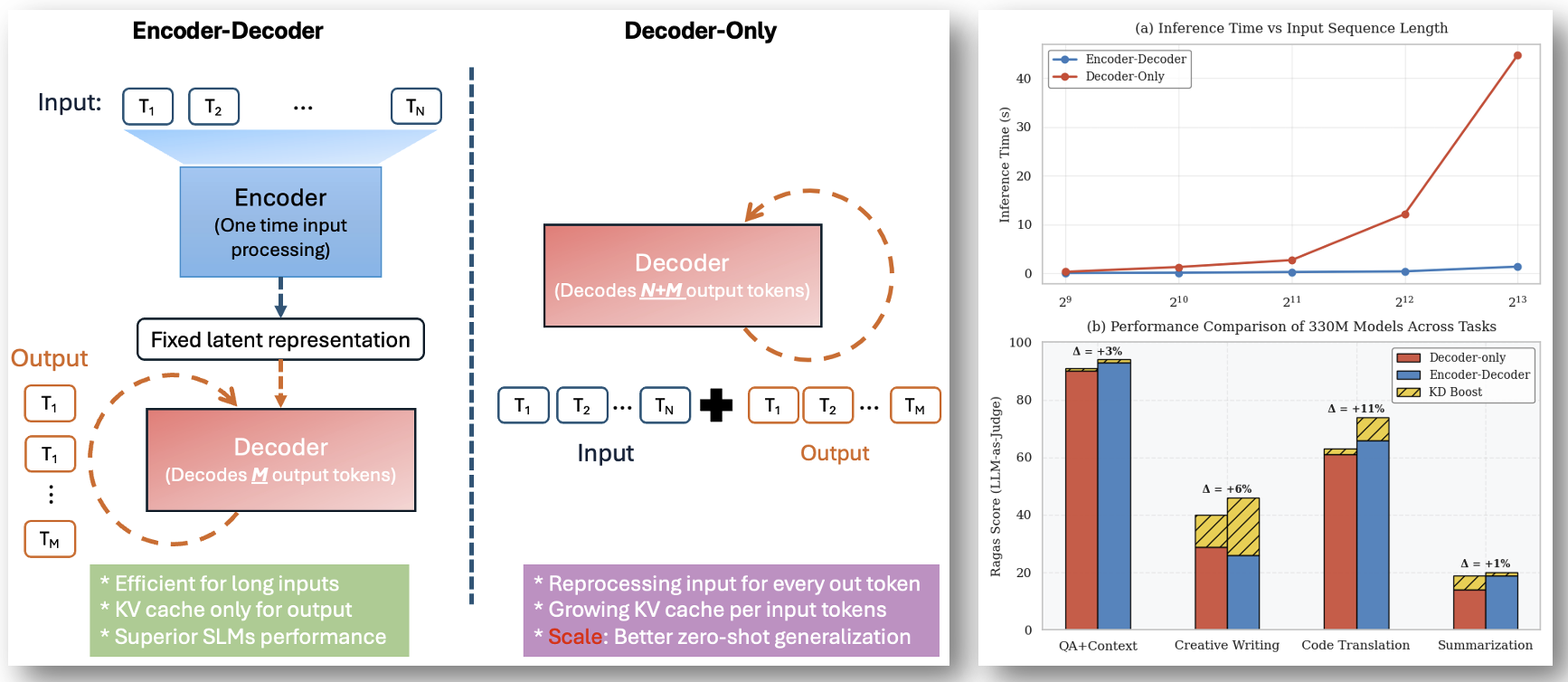}
    \caption{Architectural Efficiency in SLMs. Left: Comparison of architectures where encoder-decoder creates a fixed input representation with KV cache only for output, while decoder-only requires growing KV caches for both input and output. Top right: Inference time scaling with input length, showing encoder-decoder's efficient fixed-representation approach versus decoder-only's steeper computational growth. Bottom right: Performance across tasks showing encoder-decoder's advantages at fixed compute budget, further enhanced by KD.}
    \vspace{-10pt}
    \label{fig:architectural_comparison}
\end{figure*}
The introduction of the Transformer \cite{attention} as an encoder-decoder architecture revolutionized sequence modeling and proven remarkably robust over time. While encoder-decoder transformers demonstrated strong performance, their inherent information bottleneck between encoder and decoder components ultimately constrained further scaling to hundreds of billions of parameters, as demonstrated in \citet{t5,palm1}.

This limitation led the field toward decoder-only architectures, following Sutton's ``bitter lesson'' \cite{Sutton2019} that general methods leveraging computation often outperform specialized architectures. Models like GPT \cite{gpt} and LLaMA \cite{llama} have demonstrated impressive scaling properties \cite{scaling_laws}, achieving state-of-the-art performance through increased parameter counts by eliminating this bottleneck and enabling more flexible parameter utilization at scale.

Despite this trend, encoder-decoder architectures offer specific advantages through their separation of comprehension and generation stages. The encoder constructs a fixed representation of the input sequence, while the decoder performs targeted attention over encoded information during generation. This separation creates two key benefits: efficient handling of divergent input-output distributions (as in summarization and translation) and elimination of key-value (KV) cache requirements for input sequences, making them particularly efficient for tasks involving large inputs like code or long document reasoning/QA.

Recent advances in positional encoding, particularly Rotary Positional Embeddings (RoPE), further enhances encoder-decoder architectures in two ways: by enabling precise encoding of relative token distances for improved reasoning over long contexts, and by facilitating efficient parameter utilization through \textit{asymmetric scaling}. This dual benefit allows the architecture to both handle complex positional reasoning tasks like QA while enabling flexible optimization of encoder and decoder sizes for different tasks. Through pipelined inference, a 2N-parameter encoder-decoder model can match the computational efficiency of an N-parameter decoder-only model \cite{langarch}. 

While the architecture's inherent information bottleneck may constrain scaling to extremely large sizes \cite{t5}, this same constraint provides valuable inductive biases at more practical scales—as demonstrated by \citet{instructeval,t5}, where T5 models at just 20B parameters consistently outperform much larger architectures like the 62B parameter PaLM-1 \cite{palm1} on specific downstream tasks requiring precise positional understanding and efficient parameter utilization.

\titlespacing{\paragraph}{0pt}{0.5ex}{0.5ex}  

\begin{paragraph}{Contributions}
\begin{itemize}[leftmargin=*, noitemsep, topsep=0pt]
    \item A systematic analysis demonstrating that encoder-decoder architectures achieve superior performance (+2-4\% improvement) and efficiency (47\% lower latency) at small scales ($\leq$1B parameters) across GPU, CPU, and NPU platforms.
    
    \item A novel knowledge distillation framework enabling small encoder-decoder models to leverage capabilities from large scalable decoder-only teachers while preserving their architectural efficiency advantages.
    
    \item An extended architecture incorporating modern advances (RoPE, ViT) that maintains efficiency benefits while generalizing to vision-language tasks, showing consistent improvements in multimodal reasoning.
\end{itemize}
\end{paragraph}

\section{Related Work}
Language model architectures have evolved through the interplay of theoretical advances and practical constraints, particularly in resource-constrained deployments of small-scale language models (SLMs).

\subsection{Evolution of Transformer Architectures}
The Transformer architecture \cite{attention} revolutionized sequence modeling by replacing recurrent networks with attention mechanisms. Its core design demonstrated remarkable stability, requiring only minimal refinements such as pre-layer normalization \cite{prelayernorm} and improved positional embeddings \cite{rope}. A significant empirical study by \cite{langarch} later validated the advantages of non-causal visibility and masked language modeling, showing encoder-decoder setups consistently outperforming decoder-only architectures.

\cite{mobilellm} demonstrated that architectural choices become crucial at smaller scales. In particular, their findings revealed that deeper, thinner models and grouped-query attention (GQA) mechanisms are particularly effective for sub-billion parameter models. Building on these insights, our architecture incorporates GQA and emphasizes depth over width.

\subsection{Efficient Model Design}
Recent approaches to efficient modeling have evolved along several paths. Decoder-only solutions like SmolLM \cite{smollm} and MobileLM \cite{mobilellm} focus on higher quality data and architectural optimizations respectively, while Llama 3.2 \cite{llama3} explores pruning and knowledge distillation. While large models rely heavily on prompt engineering for task adaptation, our work suggests that finetuning smaller, more efficient models may offer a more practical and scalable approach for deployment scenarios. This hypothesis is particularly relevant for resource-constrained environments where the computational overhead of prompt engineering large models may outweigh their performance benefits.

Our study isolates architectural choice as the key variable by maintaining consistent training data, parameter counts, and optimization strategies across variants. This controlled comparison reveals fundamental differences in training objectives: decoder-only models employ uniform autoregressive objectives, while encoder-decoder architectures like T5 \cite{t5} leverage specialized objectives optimized for their architecture. Studies have shown this specialization to be better \cite{langarch,upalm}, with modern approaches often combining strategies \cite{ul2, palm2} to balance efficiency and flexibility.

Paralleling our work, \cite{modern_bert} recently showed significant improvements in encoder-only architectures through modernization of both data and architecture. Our work extends this modernization to encoder-decoder architectures while demonstrating their superiority for SLMs.

\subsection{Architectural Trade-offs}
While decoder-only models excel in zero-shot scenarios \cite{langarch}, they struggle with long-sequence generation \cite{deconly_vs_encdec} and resource utilization. Encoder-decoder architectures, in contrast, offer significant advantages through their architectural design. The one-time input processing with fixed latent representations enables substantial inference optimization, while the natural separation of phases (i.e., encoder for understanding and decoder for generation) allows for efficient parameter distribution across the model. This architecture also enables task-specific optimization through independent sizing of encoder and decoder components, providing flexible asymmetric scaling capabilities. Furthermore, the bidirectional attention mechanism delivers superior performance at smaller scales \cite{langarch}, opening opportunities for advanced optimizations such as pooling and linear attention techniques \cite{charformer}. While these benefits traditionally came with engineering challenges in handling variable-length inputs, our work shows that modern approaches like RoPE \cite{rope} can effectively mitigate these limitations while preserving the core advantages of the encoder-decoder.

\subsection{Knowledge Distillation Advances}
Knowledge distillation has emerged as a crucial bridge between efficiency and performance \cite{distillation, llama3, nvidia_llm_distillation}, with notable successes like Llama 3.3's 70B student matching its 405B teacher. Unlike previous approaches maintaining decoder-only architecture for both student and teacher, we introduce novel cross-architecture distillation, enabling small encoder-decoder models to benefit from large-scale decoder-only training while preserving their efficiency advantages.

\section{Parameter Efficient SLMs}
Building on insights from decoder-only and encoder-decoder architectures, we present a parameter-efficient framework that combines the architectural advantages of encoder-decoders with a novel knowledge distillation approach, enabling smaller models to benefit from larger decoder-only teachers while maintaining their inherent efficiency benefits.
\subsection{Architectural Design}

Our encoder-decoder architecture addresses two fundamental challenges in language modeling: efficient handling of variable-length sequences and optimal parameter allocation. While decoder-only models process concatenated input-output sequences uniformly, our approach enables specialized processing and flexible resource distribution based on the distinct roles of understanding (encoder) and generation (decoder).

\paragraph{Core Architecture} The design builds on the encoder-decoder foundation from \cite{t5}, with several key modifications for improved efficiency. Following \cite{langarch}, we maintain consistent training FLOPs across all architectural variants while incorporating modern components: pre-layer normalization \cite{prelayernorm}, Rotary Positional Embeddings (RoPE) \cite{rope}, and Grouped-Query Attention (GQA) \cite{gqa}. GQA proves particularly valuable in our resource-constrained setting, aligning with findings from \cite{mobilellm} on its effectiveness for small-scale deployments.

\paragraph{Sequence Length Management} A critical piece in our design lies in the handling of variable-length sequences. Traditional encoder-decoder architectures suffer from two key inefficiencies: separate padding requirements for encoder and decoder components, and complex cross-attention management. We address these challenges through an integrated approach to sequence processing. At the core of our solution is the integration of RoPE with neural tangent kernel (NTK) scaling \cite{palm1}, enabling flexible sequence length handling without rigid padding constraints. The encoder produces token-wise representations with fixed dimensionality (while the sequence length varies with input), enabling efficient memory utilization during generation. Furthermore, cross-attention computations leverage these pre-computed encoder representations, eliminating the need for repeated input processing. This approach maintains the architectural benefits of separate encoding and decoding while mitigating traditional efficiency bottlenecks. To validate these efficiency claims, we measured memory utilization and computational overhead when processing sequences of length 4096 with batch size 32 across both architectures. The encoder-decoder used 11-16\% less memory during inference due to its fixed-size representations, and required only 78\% of the FLOPs compared to the decoder-only model for generating 256 tokens, with the gap widening further for longer generation lengths due to the one-time input processing advantage.

\paragraph{Parameter Allocation} For our base 330M parameter models, we explore three principal encoder-decoder configurations (Table ~\ref{tab:task_performance}). The first uses a 1/3-2/3 split with 12 encoder and 32 decoder layers, the second employs a balanced 1/2-1/2 split with 22 layers each, and the third implements a 2/3-1/3 split with 32 encoder and 12 decoder layers. For comparison, our decoder-only baseline uses 48 layers to maintain parameter parity - the encoder-decoder variants use fewer total layers (44) but achieve parameter matching through additional cross-attention components in their decoder layers. Causal masking is applied throughout decoder-only variants but only in the decoder for encoder-decoder models. We find the 2/3-1/3 configuration consistently outperforms other splits, which we attribute to the inherent asymmetry in sequence-to-sequence tasks: the encoder must capture complex bidirectional relationships and long-range dependencies in the input, while the decoder's task of conditional generation can leverage the rich encoded representations and requires fewer layers for effective autoregressive modeling. This aligns with findings from attention pattern analysis showing that early layers predominantly focus on building robust contextual representations, while later layers specialize in task-specific generation \cite{t5}.

\paragraph{Scaling Analysis} We extend these configurations to 500M and 1B parameters, focusing on the best-performing 2/3-1/3 split for evaluation on SQuAD and XSum (Figure \ref{fig:scaling}). While we limit our experiments to 1B scale, understanding the transition point where encoder-decoder advantages diminish could provide valuable insights into fundamental architectural trade-offs. We encourage the research community to investigate these limits at larger scales.

Our architecture's handling of variable-length sequences becomes increasingly advantageous at scale. As sequence lengths grow, the encoder-decoder architecture provides significant efficiency gains through two mechanisms: (1) one-time input processing with storage of only final layer representations, versus decoder-only models' need to maintain KV cache states across all layers, and (2) more efficient memory utilization during generation since cross-attention operates on pre-computed encoder states. These benefits are particularly pronounced in tasks involving long documents or multi-step reasoning. The combination of efficient sequence handling and flexible parameter allocation enables our architecture to maintain strong performance while reducing computational overhead, especially for longer sequences where traditional architectures struggle with memory and computation constraints.

\subsection{Knowledge Distillation Framework}\label{sec:text_kd_align}

A key contribution of our work is a novel knowledge distillation framework that enables encoder-decoder models to learn from larger decoder-only architectures, despite their fundamentally different input-output schemas—decoder-only models use unified attention on concatenated sequences, while encoder-decoder models maintain separate processing stages.

Building upon~\cite{onpolicy}, we introduce novel sequence alignment strategies specifically designed for distilling encoder-decoder models from decoder-only teachers. For an input sequence $x$ of length $|x|$, we first generate output sequence $y$ using the student model. We structure the inputs distinctly: the teacher receives a concatenated sequence $[PAD]^{n_e} \circ x \circ y \circ [PAD]^{n_d}$ (where $[PAD]$ denotes padding tokens, $n_e$ and $n_d$ are encoder and decoder padding lengths), while the student model processes $x \circ [PAD]^{n_e}$ in the encoder and $[BOS] \circ y \circ [PAD]^{n_d}$ in the decoder (where $[BOS]$ denotes the beginning-of-sequence token). This arrangement ensures proper alignment through careful offset management, with teacher logits starting at position $(|x| + n_e - 1)$ and student logits coming directly from the decoder output.

The complete distillation process follows Algorithm \ref{alg:kd}, employing a temperature parameter $\tau$ and combining reverse KL-divergence with cross-entropy loss, where the mixing ratio is tuned per dataset. The temperature parameter serves dual purposes: softening probability distributions for improved knowledge transfer and scaling gradients through the $\tau^2$ term. Further ablation studies (see Appendix) demonstrate the effectiveness of this approach compared to alternative distillation methods.

\begin{algorithm}[t]
   \caption{Knowledge Distillation: Decoder-only to Encoder-decoder}
   \label{alg:kd}
\begin{algorithmic}[1]
   \REQUIRE Teacher model $T$, Student model $S$, Temperature $\tau$, Loss ratio $\alpha$
   \STATE {\bfseries Input:} Training batch $x$
   \STATE $y \gets \text{Generate}(S, x)$ \COMMENT{Generate with student}
   \STATE $n_e \gets \text{encoder\_seq\_len} - |x|$ \COMMENT{Encoder padding}
   \STATE $n_d \gets \text{decoder\_seq\_len} - |y|$ \COMMENT{Decoder padding}
   \STATE // Prepare sequences
   \STATE $x_t \gets [PAD]^{n_e} \circ x \circ y \circ [PAD]^{n_d}$ \COMMENT{Teacher input}
   \STATE $x_s \gets x \circ [PAD]^{n_e}$ \COMMENT{Student encoder input}
   \STATE $y_s \gets [BOS] \circ y \circ [PAD]^{n_d}$ \COMMENT{Student decoder input}
   \STATE // Forward passes
   \STATE $l_t \gets T(x_t)$ \COMMENT{Teacher logits}
   \STATE $l_s \gets S(x_s, y_s)$ \COMMENT{Student logits}
   \STATE // Align and scale logits
   \STATE $l_t \gets l_t[|x| + n_e - 1: |x| + n_e - 1 + |y|]$
   \STATE $l_s \gets l_s[:|y|]$
   \STATE $p_t \gets \text{softmax}(l_t/\tau)$
   \STATE $p_s \gets \text{softmax}(l_s/\tau)$
   \STATE // Compute mixed loss and optimize
   \STATE $L_{kd} \gets \alpha \cdot \tau^2 \cdot \text{KL}(p_t || p_s) + (1-\alpha) \cdot \text{CE}(y, y_s)$
   \STATE $S \gets \text{Optimize}(S, L_{kd})$
\end{algorithmic}
\end{algorithm}

\subsection{Training and Evaluation Methodology}

\paragraph{Training Process} Training is done in two stages. We first conduct pretraining using span corruption \citep{t5} with a 15\% noise ratio and span length $k=3$ on a decontaminated 100B token dataset from FineWeb-Edu \citep{fineweb}. For downstream tasks, we implement two training strategies: standard sequence-to-sequence learning with cross-entropy loss and knowledge distillation from a Phi-3.5Mini (3.3B parameters) \citep{phi3} that is fine-tuned on the downstream task.

\paragraph{Training Efficiency} We train all the models using 16 A100 GPUs. The encoder-decoder (2/3-1/3) model completed pretraining in 250 hours compared to 350 hours for the decoder-only variant—a mere 71\% training time while achieving a superior final performance. While decoder-only models require a quadratic attention computations over concatenated input-output sequences, encoder-decoder models process input sequences once, creating reusable fixed representations. This advantage amplifies during training where multiple output tokens are generated and evaluated for each input sequence. Our encoder-decoder model uses only 42\% of the FLOPs at 1024-token inputs compared to its decoder-only counterpart. The efficiency gap widens with scale—at 4096 tokens, computational requirements differ by 3.2x, demonstrating how one-time input processing becomes increasingly valuable with longer sequences.

\paragraph{Implementation Details} We utilize mixed precision training (BF16) with the Muon optimizer \citep{muon}, employing a learning rate of 3e-4 with 2000 warmup steps and cosine decay schedule. Our training setup processes batches of 32 samples per GPU using gradient accumulation, with maximum sequence lengths of 1024 and 256 tokens for input and output, respectively.

\paragraph{Evaluation Framework} Our evaluation framework targets tasks that stress specific computational aspects: reasoning through SQuAD-v2.0 (\cite{squad}), divergent length distributions via summarization (XSUM) (\cite{xsum}), structural transformation through code translation with CodeXGLUE (\cite{code_x_glue}), and creative generation via IELTS creative writing (\cite{ielts}). We employ two complementary metrics: Rouge-L(RL) for lexical similarity assessment and Ragas-GPT (RG) \cite{ragas} for GPT-4 based evaluation (LLM-as-a-judge) of context precision/recall, response relevancy, and output faithfulness.

\paragraph{Pretraining Results} We conducted pretraining on 100B tokens randomly sampled from FineWeb-Edu \citep{fineweb}, ensuring decontamination from evaluation sets following \citep{tulu}. After one epoch, all model variants (both encoder-decoder and decoder-only) achieved comparable performance on preplexity ($2.81\pm0.02$), and on closed-form knowledge and common-sense evaluations: MMLU ($24.82\pm0.44$), Arc-easy ($24.92\pm0.51$), and Arc-challenge ($22.95\pm0.36$), showing no statistically significant differences.

\begin{table}[t]
\centering
\scriptsize
\begin{tabularx}{\columnwidth}{lXXXX|X}
\hline
\makecell{\textbf{Model}} & \makecell{\textbf{SQuAD}\\(RL/RG)} & \makecell{\textbf{IELTS}\\(RL/RG)} & \makecell{\textbf{CodeX}\\(RL/RG)} & \makecell{\textbf{XSum}\\(RL/RG)} & \makecell{\textbf{Average}\\(RL/RG)} \\
\hline
Phi-3.3B      & 0.85/0.96          & 0.31/0.70      & 0.93/0.74          & 0.36/0.28     & 0.61/0.67 \\
\hline
\multicolumn{6}{c}{Decoder-only} \\
Seq2Seq        & 0.55/0.90          & 0.30/0.29      & 0.91/0.61          & 0.19/0.14     & 0.49/0.49 \\
KD             & 0.57/0.90          & 0.31/0.40      & 0.93/0.63          & 0.24/0.19     & 0.51/0.53 \\
\hline
\multicolumn{6}{c}{Encoder-Decoder (Seq2Seq)} \\
1/3-2/3        & 0.61/0.91          & 0.31/0.29      & 0.93/0.65          & 0.23/0.15     & 0.52/0.50 \\
1/2-1/2        & 0.64/0.92          & 0.31/0.28      & 0.93/0.66          & 0.24/0.17     & 0.53/0.51 \\
2/3-1/3        & 0.67/0.93          & 0.31/0.26      & 0.93/0.66          & 0.25/0.19     & 0.54/0.51 \\
\hline
\multicolumn{6}{c}{Encoder-Decoder (KD)} \\
1/3-2/3        & 0.62/0.93          & 0.32/0.40      & 0.93/0.70          & 0.26/0.19     & 0.53/0.56 \\
1/2-1/2        & 0.60/0.91          & 0.32/0.45      & 0.93/0.73          & 0.27/0.20     & 0.53/0.57 \\
2/3-1/3        & 0.69/0.94 & 0.32/0.46 & 0.93/0.74 & 0.27/0.20 & \textbf{0.55/0.59} \\
\hline
\end{tabularx}
\caption{Downstream Task Performance of 330M Model Variants with Varying Encoder-Decoder Allocations and Post-Training.}
\vspace{-10pt}
\label{tab:task_performance}
\end{table}

\paragraph{Downstream Performance} The empirical results presented in Table \ref{tab:task_performance} strongly validate our architectural thesis, particularly at the 330M parameter scale. The 2/3-1/3 encoder-decoder configuration with knowledge distillation demonstrates superior performance across all tasks: SQuAD 2.0 (0.69/0.94), IELTS (0.32/0.46), CodeXGLUE (0.93/0.74), and XSum (0.27/0.20). These results outperform both Seq2Seq and KD variants of the decoder-only baseline in terms of lexical overlap (RL) and reasoning capabilities (RG). These findings align with observations from \cite{flan_t5,langarch} that encoder-decoder advantages over decoder-only primarily emerge during post-training rather than pretraining.

The effectiveness of the 2/3-1/3 split, which dedicates more capacity to input processing through the encoder, supports our hypothesis regarding the importance of comprehensive input analysis in smaller models. The consistent improvements from Seq2Seq to KD variants ($+2$ - $8$ Ragas points across tasks) demonstrate that encoder-decoder architectures effectively leverage knowledge from larger models while maintaining their architectural advantages. This is particularly evident in tasks involving divergent input-output distributions (XSum: 0.27/0.20 vs 0.24/0.19) and comprehensive reasoning (SQuAD: 0.69/0.94 vs 0.57/0.90). Additionally, on standard SQuAD~2.0 F1/EM metrics, the decoder-only KD model measures 70/66 while the encoder-decoder (2/3-1/3) KD reaches 78/74. Another important remark is that while decoder-only (Seq2Seq) initially shows stronger performance in creative writing tasks (IELTS) compared to encoder-decoder (Seq2Seq) variants due to its enhanced generative capabilities and zero-shot generalization, the encoder-decoder architecture with knowledge distillation ultimately achieves superior performance (0.32/0.46 vs 0.31/0.40) even in this generative domain.

\begin{figure}[t]
\centering
\includegraphics[width=0.48\textwidth]{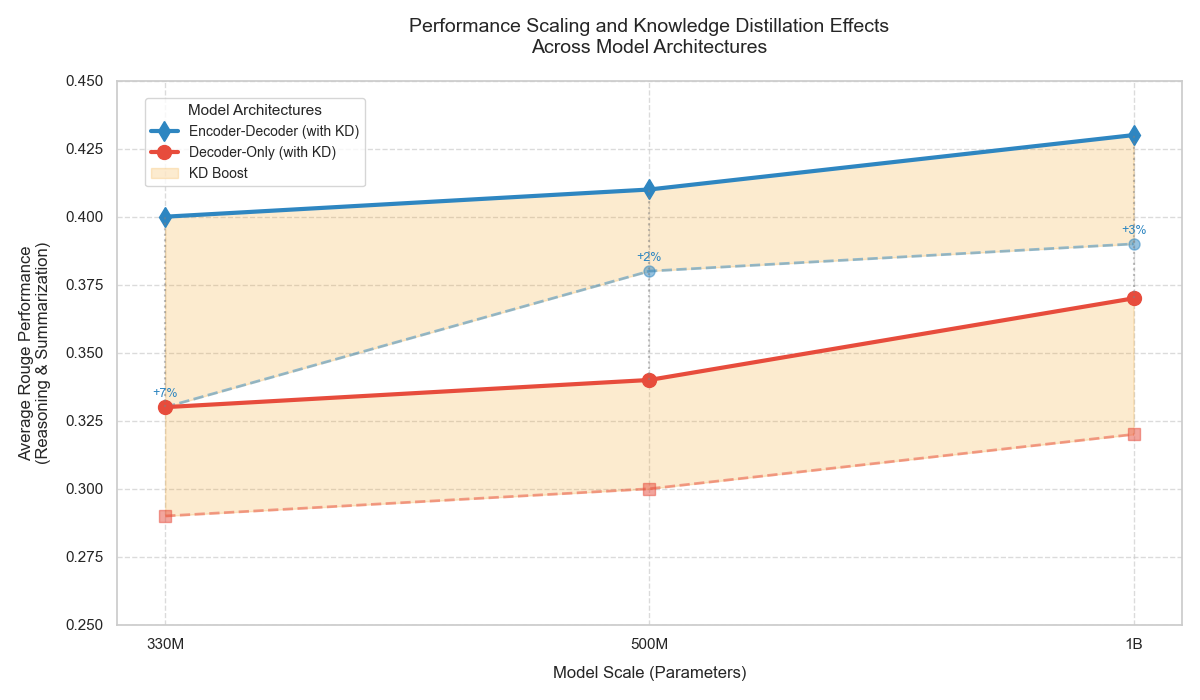}
\caption{Performance across various model scales across top two architectures (2/3-1/3 enc-dec vs dec-only).}
\vspace{-10pt}
\label{fig:scaling}
\end{figure}

\paragraph{Scaling Analysis} As illustrated in Figure \ref{fig:scaling}, the architectural advantages persist across different model scales. The performance gap between architectures remains substantial, ranging from $+7\%$ at 330M to $+6\%$ at 1B parameters. The knowledge distillation boost shows its strongest effect for encoder-decoder models at the 330M scale (+7\%), where architectural inductive biases most significantly influence the learning process. Notably, decoder-only models with KD struggle to match the performance of base encoder-decoder models at 330M and fall further behind at larger scales—at 1B parameters, decoder-only with KD underperforms even the base encoder-decoder without KD (0.37 vs. 0.39).

\subsection{Hardware Efficiency Analysis}

To demonstrate practical deployment capabilities, we conduct comprehensive efficiency analysis across three representative platforms: an NVIDIA RTX A6000-48GB GPU for high-performance computing, a X1E80100-Snapdragon Qualcomm Oryon CPU for consumer-grade client-side deployment, and a Qualcomm SnapDragon Hexagon neural processing unit (NPU) for mobile/edge applications. To maintain consistency in model execution environments, we compiled both 330M architectures into the ONNX format and executed them across CPU, GPU, and NPU platforms. Our analysis focuses on two critical metrics for inference: first token latency for response time assessment, and subsequent token speed for throughput evaluation. We report the average values over 100 examples using an input context length $512$ and a max output token length $128$. 

The results in Table \ref{tab:hardware_efficiency} demonstrate the clear efficiency advantage of encoder-decoder architectures over decoder-only models across all tested hardware platforms. Encoder-decoder models consistently achieve lower first-token latency (42\% reduction on GPU, 29\% on CPU, 47\% on NPU) and significantly higher decoding speed (3.9× on GPU, 3.8× on CPU, 4.7× on NPU). This superiority stems from the encoder-decoder’s one-time input processing, which eliminates redundant computation present in decoder-only models, making it a more efficient choice for resource-constrained deployments.

\begin{table}[t]
\centering
\footnotesize
\begin{tabular}{llcc}
\hline
\textbf{Platform} & \textbf{Model} & \textbf{First} & \textbf{Throughput} \\
& & \textbf{(ms)} & \textbf{(tok/s)} \\
\hline
GPU & Dec-only & 149 & 9.7\\
& Enc-dec (2/3-1/3) & 86 & 37.4 \\
\hline
CPU & Dec-only & 2242 & 4.0 \\
& Enc-dec (2/3-1/3) & 1591 & 15.3 \\
\hline
NPU & Dec-only & 358 & 26.5 \\
& Enc-dec (2/3-1/3) & 189 & 123.8 \\
\hline
\end{tabular}
\caption{Hardware Cross-Platform Efficiency Analysis (330M).}
\vspace{-10pt}
\label{tab:hardware_efficiency}
\end{table}

\section{Extending to Vision-Language Tasks}

Having demonstrated our architecture's efficiency for text tasks, we extend these benefits to vision-language tasks where modality separation proves even more crucial for computational efficiency.

\subsection{Vision-Language Architecture}
\begin{figure}[t]
    \centering
    \includegraphics[width=0.45\textwidth]{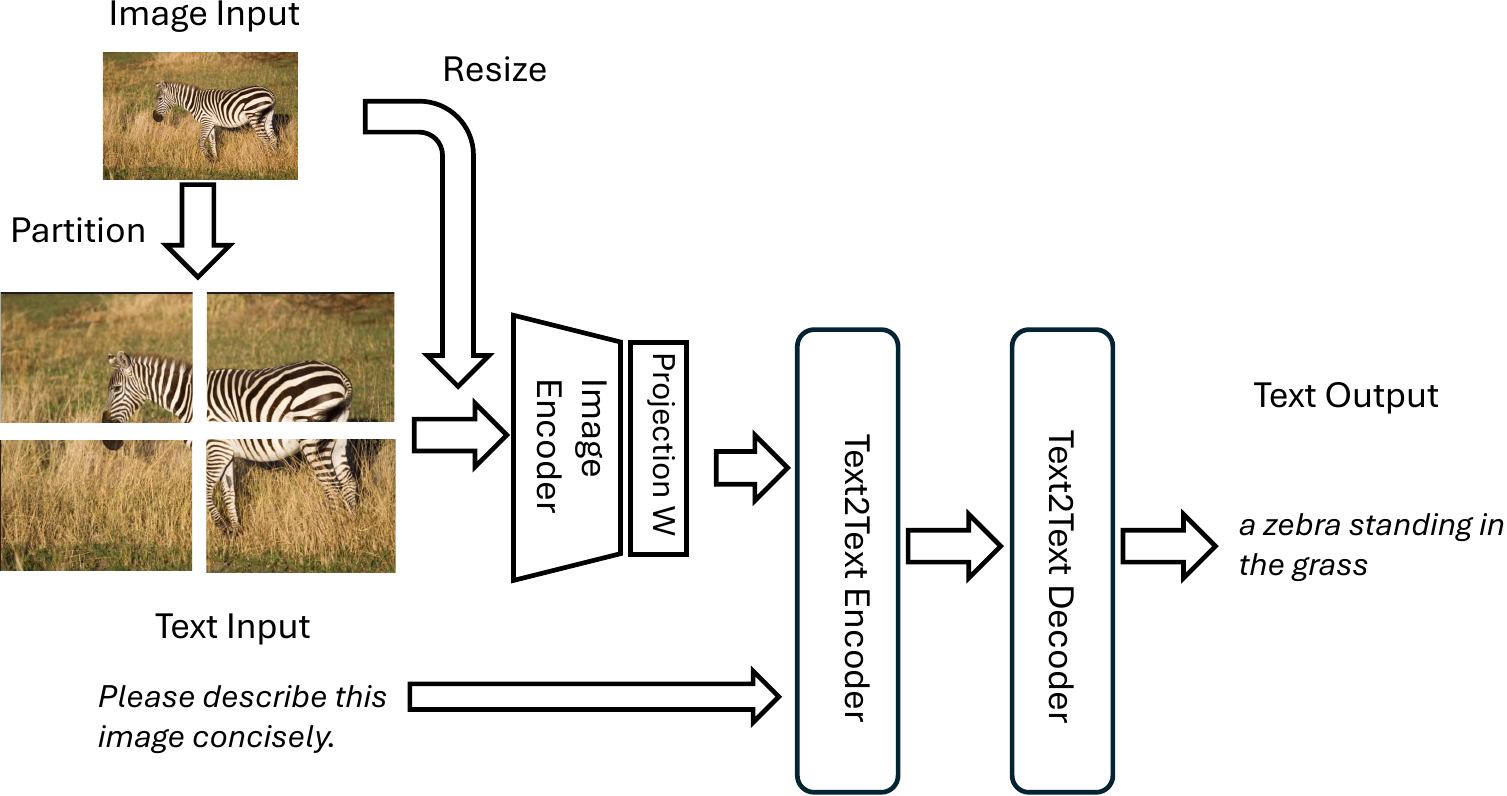}
    \caption{Vision Language Encoder-Decoder Architecture.}
    \vspace{-10pt}
    \label{fig:mm_arch_enc_dec}
\end{figure}

Our vision-language model (Figure ~\ref{fig:mm_arch_enc_dec}) maintains the core benefits of the text encoder-decoder architecture and incorporating visual processing capabilities. We utilize CLIP's ViT-L-336px~\citep{radford2021learning} as our vision encoder, choosing the highest resolution variant to maximize visual understanding. Following~\citep{liu2023visual,li2024llava}, we employ a 2-layer MLP projection layer to align the vision encoder's output with the text embedding space.

To handle high-resolution images efficiently while preserving detail, we follow \cite{li2025flexattention,chen2024dragonfly} in implementing a high-resolution image processing pipeline. Our approach first partitions input images into sub-images to enable detailed visual analysis.To provide a holistic view with a global visual context, the original image is also resized to a low-resolution thumbnail image that matches the input resolution of the vision encoder~\cite{liu2024improved}. These sub-images are then processed independently through the vision encoder, after which the encoded vision tokens are concatenated before being fed to the text encoder-decoder. This architecture maintains our principle of separation between encoding and generation, with the complete model comprising three stages: vision encoding, text encoding, and decoding. For comparison, we also implement a decoder-only variant by removing the text encoder component, allowing direct evaluation of architectural choices in the multimodal setting.

\subsection{Efficient Training Strategy}
Our process begins with feature alignment, where we train the projection layer $W$ on 600K image-caption pairs for one epoch, establishing the foundation for vision-language integration. We then enhance OCR capabilities using 200K image-OCR examples, which is critical for tasks requiring text extraction from images. This is followed by instruction following training on 700K examples to develop general visual reasoning capabilities. Finally, we fine-tune on specific downstream tasks such as VQAv2~\citep{goyal2017making} and TextVQA~\citep{singh2019towards}.

Our training leverages diverse datasets spanning multiple domains: feature alignment via LLaVA pretraining dataset~\cite{liu2023visual}, OCR understanding via UReader~\cite{ye2023ureader} and SynDoG~\cite{kim2022ocr}, diagram understanding through AI2D~\citep{kembhavi2016diagram}, chart comprehension via ChartQA~\citep{masry2022chartqa}, document analysis using DocVQA~\citep{mathew2021docvqa}, and general visual reasoning through GQA~\citep{hudson2019gqa} and VG~\citep{krishna2017visual}.

\paragraph{Vision Token Compression} Processing high-resolution images presents computational challenges, as they can generate between 5k-10k vision tokens. To address this, we implement a novel efficient token compression strategy. Our primary approach is variance-based selection, a computationally efficient heuristic that reduces vision tokens by 67\% (from 3,000 to 1,000) by identifying and filtering low-information background regions. We also explored learned token weighting through a trainable selection layer, though our experiments suggest this approach requires additional optimization through auxiliary losses.

Recent developments offer promising directions for further efficiency improvements. These include n-token concatenation as demonstrated in Leopard~\citep{jia2024leopard}, dynamic resolution selection through FlexAttention~\citep{li2025flexattention}, and semantic patch retention~\citep{chen2024dragonfly}. These methods complement our architecture and present opportunities for future optimization.

\begin{figure}[t]
\centering
\includegraphics[width=0.48\textwidth]{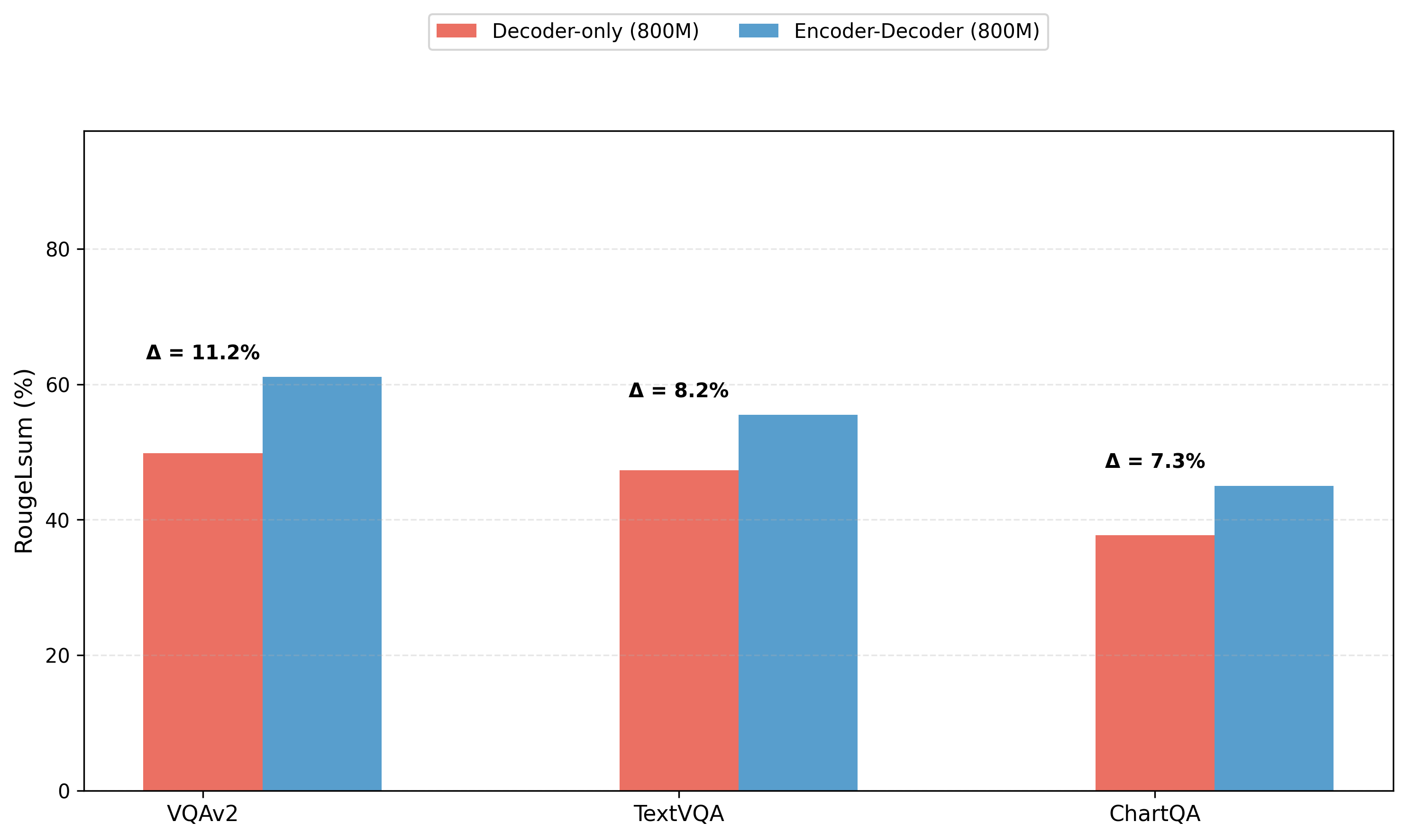}
\caption{Performance comparison across vision-language tasks. Despite equal parameter constraints (800M), our encoder-decoder architecture consistently outperforms the decoder-only baseline.}
\label{fig:vision_results}
\end{figure}

\subsection{Experimental Evaluation}
Our evaluation framework for vision-language models maintains the same principles of efficiency and effectiveness established in our text experiments. For comprehensive assessment, we compare three architectural configurations: an 1/2-1/2 encoder-decoder model with 800M parameters (22 encoder, 22 decoder layers) and a parameter-matched decoder-only baseline (48 layers). 

Our evaluation framework for vision-language models maintains the same principles of efficiency and effectiveness established in our text experiments. Building on insights from work like ~\citet{distillation} and ~\citet{onpolicy}, we leverage knowledge distillation to maximize model capabilities within our computational budget. Our evaluation compares two configurations: an encoder-decoder model with 800M parameters (22 encoder, 22 decoder layers), a similarly-distilled decoder-only baseline of 800M parameters (48 layers). Both are distilled from a task-finetuned Phi-3-Vision (4.1B parameters)~\citep{phi3} as a larger-scale reference point (teacher). We use knowledge distillation and the same training pipelines consistently across the encoder-decoder and decoder-only variants to ensure a fair comparison that isolates architectural differences while maintaining optimal performance for each configuration.

The evaluation spans several key tasks strategically chosen to stress different aspects of multimodal understanding: VQAv2~\citep{goyal2017making} for complex visual reasoning, TextVQA~\citep{singh2019towards} for cross-modal comprehension, and ChartQA~\citep{masry2022chartqa} for structured visual analysis. These tasks specifically probe the architectures' abilities to handle divergent input-output distributions and complex cross-modal reasoning. The empirical results in Figure \ref{fig:vision_results} strongly validate our architectural thesis, with the encoder-decoder model consistently outperforming its decoder-only counterpart across all vision-language tasks. Most notably, on tasks requiring complex visual reasoning and cross-modal integration (VQAv2: +11.21\%, TextVQA: +8.17\%), the encoder-decoder architecture demonstrates superior capability in handling divergent input-output distributions. The performance gap is particularly pronounced in structured visual analysis (ChartQA: +7.28\%), where the architecture's ability to maintain separate encoding and generation stages proves advantageous. While still showing headroom compared to the larger Phi-3V model, our 800M parameter encoder-decoder achieves these gains while maintaining significant parameter efficiency, reinforcing our core argument about the importance of architectural choices in resource-constrained deployments. Further ablations (see Appendix) demonstrate the effectiveness of this approach.

\section{Conclusion and discussion}
Rather than focusing on state-of-the-art comparisons with orthogonal SLMs of different training configurations like \citet{openelm} and \citet{smollm}, our work focuses on the fundamental architectural advantages of encoder-decoder designs in resource-constrained deployments. Recent trends have favored massive decoder-only models following \citet{Sutton2019}'s "bitter lesson", but our systematic investigation reveals that architectural choices become increasingly crucial as parameter budgets decrease. Intriguingly, we find that the encoder-decoder information bottleneck - often cited as a limitation for scaling to hundreds of billions of parameters - becomes a valuable inductive bias at smaller scales, effectively constraining the model to learn more efficient representations and processing patterns.

The encoder-decoder architecture delivers four key benefits in small-scale deployments:
\begin{itemize}[noitemsep, topsep=0pt, parsep=0pt]
    \item Optimized performance for asymmetric tasks and divergent input-output distributions (e.g., summarization, long-context QA)
    \item Flexible parameter allocation between components for task-specific optimization (i.e., asymmetric scaling)
    \item Training \& Inference efficiency through one-time input processing and fixed memory footprint
    \item Successful integration of modern advances like ViT \citep{vit} and RoPE \citep{rope}, extending benefits to multimodal tasks
\end{itemize}

Further, our knowledge distillation framework effectively bridges large-scale training benefits with efficient deployment, enabling smaller encoder-decoder models to leverage capabilities from larger decoder-only teachers while preserving their architectural advantages.

\paragraph{Encoder-Decoder for Reasoning Models} The emergence of advanced reasoning models like OpenAI's o1 and o3\citep{o1}, along with successful distillations like o-mini series, DeepSeek-R1, and Kimi-k1.5\citep{k15}, demonstrates the effectiveness of knowledge distillation. For smaller-scale deployments, encoder-decoder architectures offer key advantages while maintaining ability to learn from larger teachers:
\begin{itemize}[noitemsep, topsep=0pt, parsep=0pt]
    \item \textbf{Efficient Long-form Generation:} Modern applications increasingly demand lengthy generations, making per-token generation cost crucial. Encoder-decoder's one-time input processing eliminates the need for repeated computation of cached input representations
    \item \textbf{Memory Optimization:} Fixed memory footprint after initial encoding enables more efficient resource utilization compared to decoder-only models' expanding KV cache
    \item \textbf{Natural Handling of I/O Asymmetry:} The architecture's separation of understanding and generation phases aligns perfectly with modern tasks requiring extensive outputs from concise inputs (e.g., short prompts generating extensive code or detailed CoTs)
\end{itemize}

\paragraph{Future Research Directions}
While our work shows clear advantages at smaller scales, previous research \citep{t5,deconly_vs_encdec} identifies the encoder-decoder information bottleneck as a key challenge. This points to two critical areas requiring further investigation. First, we need to determine the precise scale at which this bottleneck becomes prohibitive and decoder-only architectures become more advantageous. Second, we should explore novel mechanisms for information flow between encoders and decoders, such as residual connections, to overcome scaling limitations. Additionally, developing specialized knowledge distillation techniques could allow us to combine the benefits of scalable decoder-only training with the efficient inference offered by encoder-decoder architectures.

Our findings provide concrete evidence that thoughtful architecture design can outperform scaled-down versions of larger models in resource-constrained deployments ($\leq$1B parameters). Encoder-decoder architectures with knowledge distillation demonstrate consistent improvements across tasks while reducing latency by 47\% and achieving 4.7x higher throughput. These results suggest that architectural deliberation, rather than parameter abundance, may be the key to efficient SLMs for edge and small-scale deployment.

\ifdefined\isaccepted
    \section*{Acknowledgements}
    We would like to thank Joshua Elsdon for his crucial contribution in deploying and testing our models on NPU platforms, which enabled our cross-hardware analysis. We also thank Xiaoyan Hu and Justin Wagle for their valuable insights and constructive discussions that helped shape this work.
    
    \section*{Supplementary Details}
     The following supplementary section provide detailed ablation studies and experimental analyses that further validate these architectural advantages. We begin by examining different knowledge distillation configurations, followed by exploring various approaches to vision token compression - both of which were crucial design decisions that enabled our architecture to achieve its efficiency gains while preserving model capability.
 
\subsection{Knowledge Distillation Ablation Studies}
\label{sec:kd_ablations}

\paragraph{KD Loss Function Analysis}
We conducted extensive experiments to analyze different knowledge distillation approaches on our encoder-decoder architecture, particularly focusing on the 2/3-1/3 split configuration with the SQuAD dataset. Figure \ref{fig:kd_loss_study} presents the results across different KD variants and model scales.

The study explores several key dimensions:
\begin{itemize}
    \item \textbf{Loss Mixing Parameter ($\alpha$)}: We varied $\alpha$ from 0.0 to 1.0, where $\alpha=0$ corresponds to pure sequence-to-sequence learning and $\alpha=1.0$ represents pure KD loss.
    \item \textbf{Teacher-Student Generation}: We compared forward teacher (using teacher generations), reverse teacher (using teacher generations), and forward student (using student generations for KD).
    \item \textbf{Model Scaling}: We evaluated the same KD method across different model sizes (330M, 500M, and compared against Phi3.5).
\end{itemize}

\begin{figure}[t]
\centering
\includegraphics[width=0.48\textwidth]{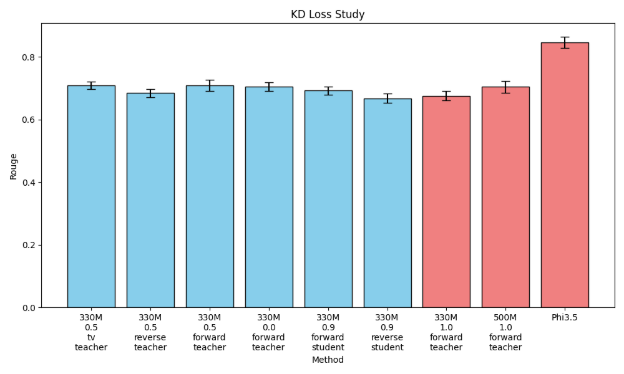}
\caption{Knowledge Distillation ablation study on SQuAD showing Rouge-L scores across different KD configurations. First six columns compare KD methods (forward/reverse) and loss mixing ratios ($\alpha$) on 330M models, where $\alpha=0.0$ represents pure sequence-to-sequence learning. Last three columns demonstrate performance scaling from 330M to Phi3.5. Error bars show standard deviation across 3 runs.}
\label{fig:kd_loss_study}
\end{figure}

Key findings from this ablation study include:
\begin{enumerate}
    \item For SQuAD specifically, sequence-to-sequence learning ($\alpha=0.0$) shows strong performance, with results generally improving as $\alpha$ approaches 0. This trend was unique to SQuAD and not observed across other datasets.
    
    \item Different KD loss functions (forward vs. reverse) show comparable performance, suggesting that for this specific task and architecture, the choice of KD method is less critical than other hyperparameters.
    
    \item Using student generations for KD (on-policy distillation) demonstrates competitive performance while offering practical advantages: faster training times and elimination of teacher generation caching requirements.
    
    \item The scaling behavior across model sizes (330M → 500M → Phi3.5) follows expected scaling laws, providing validation for our architectural choices at different scales.
\end{enumerate}

These findings align with recent work on generalized knowledge distillation \cite{onpolicy}, particularly regarding the effectiveness of on-policy distillation. However, our results suggest that the optimal KD strategy may be task-dependent, especially for specialized tasks like question answering.

\subsection{Vision Token Compression Analysis}
\label{sec:vision_ablations}

For our vision-language integration, we conducted detailed ablation studies on different token compression and cross-attention strategies, as shown in Table \ref{tab:ablation_vl_study}. Our experiments focused on the TextVQA dataset, which requires both strong visual understanding and text comprehension.

\begin{table}[t]
\centering
\begin{tabular}{lcccc}
\hline
\textbf{Model} & \textbf{RougeLSum} \\
\hline
Baseline (seq2seq) & 0.5506  \\
\hline
Vision-cross attn on decoder   & 0.5546                 \\
Vision-cross attn on encoder        & 0.5544                 \\
Variance-based selection             & 0.5383               \\
Learned token weighting        & 0.4787               \\

\hline
\end{tabular}
\caption{Ablation Study Across Vision-Language Model Variants on TextVQA}
\label{tab:ablation_vl_study}
\end{table}

We evaluated several architectural variants:

\begin{itemize}
    \item \textbf{Baseline}: Standard sequence-to-sequence architecture without specialized vision components
    
    \item \textbf{Cross-attention Placement}: We tested vision-cross attention in both decoder and encoder components
    
    \item \textbf{Token Selection Strategies}: We compared two approaches for managing vision tokens:
    \begin{itemize}
        \item Variance-based selection: A computationally efficient approach that reduces vision tokens by 67\% (from 3,000 to 1,000)
        \item Learned token weighting: A trainable selection mechanism using auxiliary losses
    \end{itemize}
\end{itemize}

Key findings from this study include:

\begin{enumerate}
    \item Both encoder and decoder cross-attention placements show similar improvements over the baseline (+0.004 Rouge-L), suggesting that the presence of vision-text interaction is more important than its specific location.
    
    \item Variance-based selection, while showing a small performance decrease (-0.012 Rouge-L), offers a favorable efficiency-performance trade-off given its significant reduction in computational requirements.
    
    \item Learned token weighting, despite its theoretical appeal, shows degraded performance (-0.072 Rouge-L). This may be due to optimization challenges or the complexity of training the selection mechanism jointly with the main task.
\end{enumerate}

These results informed our final architecture choices, favoring the simpler variance-based selection approach for its balance of efficiency and performance. The comparable performance of encoder and decoder cross-attention also supports our architectural principle of maintaining clean separation between modality processing stages. 
\fi

\bibliography{main}

\begin{thebibliography}{54}
\providecommand{\natexlab}[1]{#1}
\providecommand{\url}[1]{\texttt{#1}}
\expandafter\ifx\csname urlstyle\endcsname\relax
  \providecommand{\doi}[1]{doi: #1}\else
  \providecommand{\doi}{doi: \begingroup \urlstyle{rm}\Url}\fi

\bibitem[Abdin et~al.(2024)Abdin, Aneja, Awadalla, Awadallah, Awan, Bach,
  Bahree, Bakhtiari, Bao, Behl, et~al.]{phi3}
Abdin, M., Aneja, J., Awadalla, H., Awadallah, A., Awan, A.~A., Bach, N.,
  Bahree, A., Bakhtiari, A., Bao, J., Behl, H., et~al.
\newblock Phi-3 technical report: A highly capable language model locally on
  your phone.
\newblock \emph{arXiv preprint arXiv:2404.14219}, 2024.

\bibitem[Agarwal et~al.(2024)Agarwal, Vieillard, Zhou, Stanczyk, Garea, Geist,
  and Bachem]{onpolicy}
Agarwal, R., Vieillard, N., Zhou, Y., Stanczyk, P., Garea, S.~R., Geist, M.,
  and Bachem, O.
\newblock On-policy distillation of language models: Learning from
  self-generated mistakes.
\newblock In \emph{The Twelfth International Conference on Learning
  Representations}, 2024.

\bibitem[Ainslie et~al.(2023)Ainslie, Lee-Thorp, de~Jong, Zemlyanskiy, Lebrón,
  and Sanghai]{gqa}
Ainslie, J., Lee-Thorp, J., de~Jong, M., Zemlyanskiy, Y., Lebrón, F., and
  Sanghai, S.
\newblock Gqa: Training generalized multi-query transformer models from
  multi-head checkpoints, 2023.

\bibitem[Alexey(2020)]{vit}
Alexey, D.
\newblock An image is worth 16x16 words: Transformers for image recognition at
  scale.
\newblock \emph{arXiv preprint arXiv: 2010.11929}, 2020.

\bibitem[Allal et~al.(2024)Allal, Lozhkov, and Bakouch]{smollm}
Allal, L.~B., Lozhkov, A., and Bakouch, E.
\newblock Smollm - blazingly fast and remarkably powerful.
\newblock \url{https://huggingface.co/blog/smollm}, 2024.

\bibitem[Anil et~al.(2023)Anil, Dai, Firat, Johnson, Lepikhin, Passos, Shakeri,
  Taropa, Bailey, Chen, et~al.]{palm2}
Anil, R., Dai, A.~M., Firat, O., Johnson, M., Lepikhin, D., Passos, A.,
  Shakeri, S., Taropa, E., Bailey, P., Chen, Z., et~al.
\newblock Palm 2 technical report.
\newblock \emph{arXiv preprint arXiv:2305.10403}, 2023.

\bibitem[Chen et~al.(2024)Chen, Thapa, Chalamala, Athiwaratkun, Song, and
  Zou]{chen2024dragonfly}
Chen, K., Thapa, R., Chalamala, R., Athiwaratkun, B., Song, S.~L., and Zou, J.
\newblock Dragonfly: Multi-resolution zoom supercharges large visual-language
  model.
\newblock \emph{arXiv preprint arXiv:2406.00977}, 2024.

\bibitem[Chia et~al.(2023)Chia, Hong, Bing, and Poria]{instructeval}
Chia, Y.~K., Hong, P., Bing, L., and Poria, S.
\newblock Instructeval: Towards holistic evaluation of instruction-tuned large
  language models.
\newblock \emph{arXiv preprint arXiv:2306.04757}, 2023.

\bibitem[Chillies(2024)]{ielts}
Chillies.
\newblock Ielts writing task 2 evaluation dataset, 2024.
\newblock URL
  \url{https://huggingface.co/datasets/chillies/IELTS-writing-task-2-evaluation}.
\newblock Accessed: 2024-12-03.

\bibitem[Chowdhery et~al.(2023)Chowdhery, Narang, Devlin, Bosma, Mishra,
  Roberts, Barham, Chung, Sutton, Gehrmann, et~al.]{palm1}
Chowdhery, A., Narang, S., Devlin, J., Bosma, M., Mishra, G., Roberts, A.,
  Barham, P., Chung, H.~W., Sutton, C., Gehrmann, S., et~al.
\newblock Palm: Scaling language modeling with pathways.
\newblock \emph{Journal of Machine Learning Research}, 24\penalty0
  (240):\penalty0 1--113, 2023.

\bibitem[Chung et~al.(2024)Chung, Hou, Longpre, Zoph, Tay, Fedus, Li, Wang,
  Dehghani, Brahma, et~al.]{flan_t5}
Chung, H.~W., Hou, L., Longpre, S., Zoph, B., Tay, Y., Fedus, W., Li, Y., Wang,
  X., Dehghani, M., Brahma, S., et~al.
\newblock Scaling instruction-finetuned language models.
\newblock \emph{Journal of Machine Learning Research}, 25\penalty0
  (70):\penalty0 1--53, 2024.

\bibitem[Dubey et~al.(2024)Dubey, Jauhri, Pandey, Kadian, Al-Dahle, Letman,
  Mathur, Schelten, Yang, Fan, et~al.]{llama3}
Dubey, A., Jauhri, A., Pandey, A., Kadian, A., Al-Dahle, A., Letman, A.,
  Mathur, A., Schelten, A., Yang, A., Fan, A., et~al.
\newblock The llama 3 herd of models.
\newblock \emph{arXiv preprint arXiv:2407.21783}, 2024.

\bibitem[Fu et~al.(2023)Fu, Lam, Yu, So, Hu, Liu, and
  Collier]{deconly_vs_encdec}
Fu, Z., Lam, W., Yu, Q., So, A. M.-C., Hu, S., Liu, Z., and Collier, N.
\newblock Decoder-only or encoder-decoder? interpreting language model as a
  regularized encoder-decoder.
\newblock \emph{arXiv preprint arXiv:2304.04052}, 2023.

\bibitem[Goyal et~al.(2017)Goyal, Khot, Summers-Stay, Batra, and
  Parikh]{goyal2017making}
Goyal, Y., Khot, T., Summers-Stay, D., Batra, D., and Parikh, D.
\newblock Making the v in vqa matter: Elevating the role of image understanding
  in visual question answering.
\newblock In \emph{Proceedings of the IEEE conference on computer vision and
  pattern recognition}, pp.\  6904--6913, 2017.

\bibitem[Gradients(2024)]{ragas}
Gradients, E.
\newblock \emph{Ragas Documentation}, 2024.
\newblock URL \url{https://docs.ragas.io/en/stable/}.
\newblock Accessed: 2024-12-07.

\bibitem[Hinton(2015)]{distillation}
Hinton, G.
\newblock Distilling the knowledge in a neural network.
\newblock \emph{arXiv preprint arXiv:1503.02531}, 2015.

\bibitem[Hudson \& Manning(2019)Hudson and Manning]{hudson2019gqa}
Hudson, D.~A. and Manning, C.~D.
\newblock Gqa: A new dataset for real-world visual reasoning and compositional
  question answering.
\newblock In \emph{Proceedings of the IEEE/CVF conference on computer vision
  and pattern recognition}, pp.\  6700--6709, 2019.

\bibitem[Jia et~al.(2024)Jia, Yu, Ma, Fang, Zhang, Ouyang, Zhang, Jiang, and
  Yu]{jia2024leopard}
Jia, M., Yu, W., Ma, K., Fang, T., Zhang, Z., Ouyang, S., Zhang, H., Jiang, M.,
  and Yu, D.
\newblock Leopard: A vision language model for text-rich multi-image tasks.
\newblock \emph{arXiv preprint arXiv:2410.01744}, 2024.

\bibitem[Jordan et~al.(2024)Jordan, Jin, Boza, Jiacheng, Cecista, Newhouse, and
  Bernstein]{muon}
Jordan, K., Jin, Y., Boza, V., Jiacheng, Y., Cecista, F., Newhouse, L., and
  Bernstein, J.
\newblock Muon: An optimizer for hidden layers in neural networks, 2024.
\newblock URL \url{https://kellerjordan.github.io/posts/muon/}.

\bibitem[Kaplan et~al.(2020)Kaplan, McCandlish, Henighan, Brown, Chess, Child,
  Gray, Radford, Wu, and Amodei]{scaling_laws}
Kaplan, J., McCandlish, S., Henighan, T., Brown, T.~B., Chess, B., Child, R.,
  Gray, S., Radford, A., Wu, J., and Amodei, D.
\newblock Scaling laws for neural language models.
\newblock \emph{arXiv preprint arXiv:2001.08361}, 2020.

\bibitem[Kembhavi et~al.(2016)Kembhavi, Salvato, Kolve, Seo, Hajishirzi, and
  Farhadi]{kembhavi2016diagram}
Kembhavi, A., Salvato, M., Kolve, E., Seo, M., Hajishirzi, H., and Farhadi, A.
\newblock A diagram is worth a dozen images.
\newblock In \emph{Computer Vision--ECCV 2016: 14th European Conference,
  Amsterdam, The Netherlands, October 11--14, 2016, Proceedings, Part IV 14},
  pp.\  235--251. Springer, 2016.

\bibitem[Kim et~al.(2022)Kim, Hong, Yim, Nam, Park, Yim, Hwang, Yun, Han, and
  Park]{kim2022ocr}
Kim, G., Hong, T., Yim, M., Nam, J., Park, J., Yim, J., Hwang, W., Yun, S.,
  Han, D., and Park, S.
\newblock Ocr-free document understanding transformer.
\newblock In \emph{European Conference on Computer Vision}, pp.\  498--517.
  Springer, 2022.

\bibitem[Krishna et~al.(2017)Krishna, Zhu, Groth, Johnson, Hata, Kravitz, Chen,
  Kalantidis, Li, Shamma, et~al.]{krishna2017visual}
Krishna, R., Zhu, Y., Groth, O., Johnson, J., Hata, K., Kravitz, J., Chen, S.,
  Kalantidis, Y., Li, L.-J., Shamma, D.~A., et~al.
\newblock Visual genome: Connecting language and vision using crowdsourced
  dense image annotations.
\newblock \emph{International journal of computer vision}, 123:\penalty0
  32--73, 2017.

\bibitem[Lambert et~al.(2024)Lambert, Morrison, Pyatkin, Huang, Ivison,
  Brahman, Miranda, Liu, Dziri, Lyu, et~al.]{tulu}
Lambert, N., Morrison, J., Pyatkin, V., Huang, S., Ivison, H., Brahman, F.,
  Miranda, L. J.~V., Liu, A., Dziri, N., Lyu, S., et~al.
\newblock T$\backslash$" ulu 3: Pushing frontiers in open language model
  post-training.
\newblock \emph{arXiv preprint arXiv:2411.15124}, 2024.

\bibitem[Li et~al.(2024)Li, Zhang, Guo, Zhang, Li, Zhang, Zhang, Zhang, Li,
  Liu, et~al.]{li2024llava}
Li, B., Zhang, Y., Guo, D., Zhang, R., Li, F., Zhang, H., Zhang, K., Zhang, P.,
  Li, Y., Liu, Z., et~al.
\newblock Llava-onevision: Easy visual task transfer.
\newblock \emph{arXiv preprint arXiv:2408.03326}, 2024.

\bibitem[Li et~al.(2025)Li, Chen, Cai, Chen, Hong, Chen, Shen, and
  Gan]{li2025flexattention}
Li, J., Chen, D., Cai, T., Chen, P., Hong, Y., Chen, Z., Shen, Y., and Gan, C.
\newblock Flexattention for efficient high-resolution vision-language models.
\newblock In \emph{European Conference on Computer Vision}, pp.\  286--302.
  Springer, 2025.

\bibitem[Liu et~al.(2023)Liu, Li, Wu, and Lee]{liu2023visual}
Liu, H., Li, C., Wu, Q., and Lee, Y.~J.
\newblock Visual instruction tuning.
\newblock \emph{Advances in neural information processing systems}, 36, 2023.

\bibitem[Liu et~al.(2024{\natexlab{a}})Liu, Li, Li, and Lee]{liu2024improved}
Liu, H., Li, C., Li, Y., and Lee, Y.~J.
\newblock Improved baselines with visual instruction tuning.
\newblock In \emph{Proceedings of the IEEE/CVF Conference on Computer Vision
  and Pattern Recognition}, pp.\  26296--26306, 2024{\natexlab{a}}.

\bibitem[Liu et~al.(2024{\natexlab{b}})Liu, Zhao, Iandola, Lai, Tian, Fedorov,
  Xiong, Chang, Shi, Krishnamoorthi, et~al.]{mobilellm}
Liu, Z., Zhao, C., Iandola, F., Lai, C., Tian, Y., Fedorov, I., Xiong, Y.,
  Chang, E., Shi, Y., Krishnamoorthi, R., et~al.
\newblock Mobilellm: Optimizing sub-billion parameter language models for
  on-device use cases.
\newblock \emph{arXiv preprint arXiv:2402.14905}, 2024{\natexlab{b}}.

\bibitem[Lu et~al.(2021)Lu, Guo, Ren, Huang, Svyatkovskiy, Blanco, Clement,
  Drain, Jiang, Tang, Li, Zhou, Shou, Zhou, Tufano, Gong, Zhou, Duan,
  Sundaresan, Deng, Fu, and Liu]{code_x_glue}
Lu, S., Guo, D., Ren, S., Huang, J., Svyatkovskiy, A., Blanco, A., Clement,
  C.~B., Drain, D., Jiang, D., Tang, D., Li, G., Zhou, L., Shou, L., Zhou, L.,
  Tufano, M., Gong, M., Zhou, M., Duan, N., Sundaresan, N., Deng, S.~K., Fu,
  S., and Liu, S.
\newblock Codexglue: {A} machine learning benchmark dataset for code
  understanding and generation.
\newblock \emph{CoRR}, abs/2102.04664, 2021.

\bibitem[Masry et~al.(2022)Masry, Long, Tan, Joty, and Hoque]{masry2022chartqa}
Masry, A., Long, D.~X., Tan, J.~Q., Joty, S., and Hoque, E.
\newblock Chartqa: A benchmark for question answering about charts with visual
  and logical reasoning.
\newblock \emph{arXiv preprint arXiv:2203.10244}, 2022.

\bibitem[Mathew et~al.(2021)Mathew, Karatzas, and Jawahar]{mathew2021docvqa}
Mathew, M., Karatzas, D., and Jawahar, C.
\newblock Docvqa: A dataset for vqa on document images.
\newblock In \emph{Proceedings of the IEEE/CVF winter conference on
  applications of computer vision}, pp.\  2200--2209, 2021.

\bibitem[Mehta et~al.(2024)Mehta, Sekhavat, Cao, Horton, Jin, Sun, Mirzadeh,
  Najibi, Belenko, Zatloukal, et~al.]{openelm}
Mehta, S., Sekhavat, M.~H., Cao, Q., Horton, M., Jin, Y., Sun, C., Mirzadeh,
  S.~I., Najibi, M., Belenko, D., Zatloukal, P., et~al.
\newblock Openelm: An efficient language model family with open training and
  inference framework.
\newblock In \emph{Workshop on Efficient Systems for Foundation Models II@
  ICML2024}, 2024.

\bibitem[Narayan et~al.(2018)Narayan, Cohen, and Lapata]{xsum}
Narayan, S., Cohen, S.~B., and Lapata, M.
\newblock Don't give me the details, just the summary! topic-aware
  convolutional neural networks for extreme summarization.
\newblock \emph{arXiv preprint arXiv:1808.08745}, 2018.

\bibitem[OpenAI et~al.(2024)OpenAI, :, Jaech, Kalai, Lerer, Richardson,
  El-Kishky, Low, Helyar, Madry, Beutel, Carney, Iftimie, Karpenko, Passos,
  Neitz, Prokofiev, Wei, Tam, Bennett, Kumar, Saraiva, Vallone, Duberstein,
  Kondrich, Mishchenko, Applebaum, Jiang, Nair, Zoph, Ghorbani, Rossen,
  Sokolowsky, Barak, McGrew, Minaiev, Hao, Baker, Houghton, McKinzie, Eastman,
  Lugaresi, Bassin, Hudson, Li, de~Bourcy, Voss, Shen, Zhang, Koch, Orsinger,
  Hesse, Fischer, Chan, Roberts, Kappler, Levy, Selsam, Dohan, Farhi, Mely,
  Robinson, Tsipras, Li, Oprica, Freeman, Zhang, Wong, Proehl, Cheung,
  Mitchell, Wallace, Ritter, Mays, Wang, Such, Raso, Leoni, Tsimpourlas, Song,
  von Lohmann, Sulit, Salmon, Parascandolo, Chabot, Zhao, Brockman, Leclerc,
  Salman, Bao, Sheng, Andrin, Bagherinezhad, Ren, Lightman, Chung, Kivlichan,
  O'Connell, Osband, Gilaberte, Akkaya, Kostrikov, Sutskever, Kofman, Pachocki,
  Lennon, Wei, Harb, Twore, Feng, Yu, Weng, Tang, Yu, Candela, Palermo, Parish,
  Heidecke, Hallman, Rizzo, Gordon, Uesato, Ward, Huizinga, Wang, Chen, Xiao,
  Singhal, Nguyen, Cobbe, Shi, Wood, Rimbach, Gu-Lemberg, Liu, Lu, Stone, Yu,
  Ahmad, Yang, Liu, Maksin, Ho, Fedus, Weng, Li, McCallum, Held, Kuhn,
  Kondraciuk, Kaiser, Metz, Boyd, Trebacz, Joglekar, Chen, Tintor, Meyer,
  Jones, Kaufer, Schwarzer, Shah, Yatbaz, Guan, Xu, Yan, Glaese, Chen, Lampe,
  Malek, Wang, Fradin, McClay, Pavlov, Wang, Wang, Murati, Bavarian,
  Rohaninejad, McAleese, Chowdhury, Chowdhury, Ryder, Tezak, Brown, Nachum,
  Boiko, Murk, Watkins, Chao, Ashbourne, Izmailov, Zhokhov, Dias, Arora, Lin,
  Lopes, Gaon, Miyara, Leike, Hwang, Garg, Brown, James, Shu, Cheu, Greene,
  Jain, Altman, Toizer, Toyer, Miserendino, Agarwal, Hernandez, Baker,
  McKinney, Yan, Zhao, Hu, Santurkar, Chaudhuri, Zhang, Fu, Papay, Lin, Balaji,
  Sanjeev, Sidor, Broda, Clark, Wang, Gordon, Sanders, Patwardhan, Sottiaux,
  Degry, Dimson, Zheng, Garipov, Stasi, Bansal, Creech, Peterson, Eloundou, Qi,
  Kosaraju, Monaco, Pong, Fomenko, Zheng, Zhou, McCabe, Zaremba, Dubois, Lu,
  Chen, Cha, Bai, He, Zhang, Wang, Shao, and Li]{o1}
OpenAI, :, Jaech, A., Kalai, A., Lerer, A., Richardson, A., El-Kishky, A., Low,
  A., Helyar, A., Madry, A., Beutel, A., Carney, A., Iftimie, A., Karpenko, A.,
  Passos, A.~T., Neitz, A., Prokofiev, A., Wei, A., Tam, A., Bennett, A.,
  Kumar, A., Saraiva, A., Vallone, A., Duberstein, A., Kondrich, A.,
  Mishchenko, A., Applebaum, A., Jiang, A., Nair, A., Zoph, B., Ghorbani, B.,
  Rossen, B., Sokolowsky, B., Barak, B., McGrew, B., Minaiev, B., Hao, B.,
  Baker, B., Houghton, B., McKinzie, B., Eastman, B., Lugaresi, C., Bassin, C.,
  Hudson, C., Li, C.~M., de~Bourcy, C., Voss, C., Shen, C., Zhang, C., Koch,
  C., Orsinger, C., Hesse, C., Fischer, C., Chan, C., Roberts, D., Kappler, D.,
  Levy, D., Selsam, D., Dohan, D., Farhi, D., Mely, D., Robinson, D., Tsipras,
  D., Li, D., Oprica, D., Freeman, E., Zhang, E., Wong, E., Proehl, E., Cheung,
  E., Mitchell, E., Wallace, E., Ritter, E., Mays, E., Wang, F., Such, F.~P.,
  Raso, F., Leoni, F., Tsimpourlas, F., Song, F., von Lohmann, F., Sulit, F.,
  Salmon, G., Parascandolo, G., Chabot, G., Zhao, G., Brockman, G., Leclerc,
  G., Salman, H., Bao, H., Sheng, H., Andrin, H., Bagherinezhad, H., Ren, H.,
  Lightman, H., Chung, H.~W., Kivlichan, I., O'Connell, I., Osband, I.,
  Gilaberte, I.~C., Akkaya, I., Kostrikov, I., Sutskever, I., Kofman, I.,
  Pachocki, J., Lennon, J., Wei, J., Harb, J., Twore, J., Feng, J., Yu, J.,
  Weng, J., Tang, J., Yu, J., Candela, J.~Q., Palermo, J., Parish, J.,
  Heidecke, J., Hallman, J., Rizzo, J., Gordon, J., Uesato, J., Ward, J.,
  Huizinga, J., Wang, J., Chen, K., Xiao, K., Singhal, K., Nguyen, K., Cobbe,
  K., Shi, K., Wood, K., Rimbach, K., Gu-Lemberg, K., Liu, K., Lu, K., Stone,
  K., Yu, K., Ahmad, L., Yang, L., Liu, L., Maksin, L., Ho, L., Fedus, L.,
  Weng, L., Li, L., McCallum, L., Held, L., Kuhn, L., Kondraciuk, L., Kaiser,
  L., Metz, L., Boyd, M., Trebacz, M., Joglekar, M., Chen, M., Tintor, M.,
  Meyer, M., Jones, M., Kaufer, M., Schwarzer, M., Shah, M., Yatbaz, M., Guan,
  M.~Y., Xu, M., Yan, M., Glaese, M., Chen, M., Lampe, M., Malek, M., Wang, M.,
  Fradin, M., McClay, M., Pavlov, M., Wang, M., Wang, M., Murati, M., Bavarian,
  M., Rohaninejad, M., McAleese, N., Chowdhury, N., Chowdhury, N., Ryder, N.,
  Tezak, N., Brown, N., Nachum, O., Boiko, O., Murk, O., Watkins, O., Chao, P.,
  Ashbourne, P., Izmailov, P., Zhokhov, P., Dias, R., Arora, R., Lin, R.,
  Lopes, R.~G., Gaon, R., Miyara, R., Leike, R., Hwang, R., Garg, R., Brown,
  R., James, R., Shu, R., Cheu, R., Greene, R., Jain, S., Altman, S., Toizer,
  S., Toyer, S., Miserendino, S., Agarwal, S., Hernandez, S., Baker, S.,
  McKinney, S., Yan, S., Zhao, S., Hu, S., Santurkar, S., Chaudhuri, S.~R.,
  Zhang, S., Fu, S., Papay, S., Lin, S., Balaji, S., Sanjeev, S., Sidor, S.,
  Broda, T., Clark, A., Wang, T., Gordon, T., Sanders, T., Patwardhan, T.,
  Sottiaux, T., Degry, T., Dimson, T., Zheng, T., Garipov, T., Stasi, T.,
  Bansal, T., Creech, T., Peterson, T., Eloundou, T., Qi, V., Kosaraju, V.,
  Monaco, V., Pong, V., Fomenko, V., Zheng, W., Zhou, W., McCabe, W., Zaremba,
  W., Dubois, Y., Lu, Y., Chen, Y., Cha, Y., Bai, Y., He, Y., Zhang, Y., Wang,
  Y., Shao, Z., and Li, Z.
\newblock Openai o1 system card, 2024.

\bibitem[Penedo et~al.(2024)Penedo, Kydlíček, allal, Lozhkov, Mitchell,
  Raffel, Werra, and Wolf]{fineweb}
Penedo, G., Kydlíček, H., allal, L.~B., Lozhkov, A., Mitchell, M., Raffel,
  C., Werra, L.~V., and Wolf, T.
\newblock The fineweb datasets: Decanting the web for the finest text data at
  scale, 2024.
\newblock URL \url{https://arxiv.org/abs/2406.17557}.

\bibitem[Radford(2018)]{gpt}
Radford, A.
\newblock Improving language understanding by generative pre-training.
\newblock \emph{Advances in Neural Information Processing Systems (2017)},
  2018.

\bibitem[Radford et~al.(2021)Radford, Kim, Hallacy, Ramesh, Goh, Agarwal,
  Sastry, Askell, Mishkin, Clark, et~al.]{radford2021learning}
Radford, A., Kim, J.~W., Hallacy, C., Ramesh, A., Goh, G., Agarwal, S., Sastry,
  G., Askell, A., Mishkin, P., Clark, J., et~al.
\newblock Learning transferable visual models from natural language
  supervision.
\newblock In \emph{International conference on machine learning}, pp.\
  8748--8763. PMLR, 2021.

\bibitem[Raffel et~al.(2020)Raffel, Shazeer, Roberts, Lee, Narang, Matena,
  Zhou, Li, and Liu]{t5}
Raffel, C., Shazeer, N., Roberts, A., Lee, K., Narang, S., Matena, M., Zhou,
  Y., Li, W., and Liu, P.~J.
\newblock Exploring the limits of transfer learning with a unified text-to-text
  transformer.
\newblock \emph{Journal of machine learning research}, 21\penalty0
  (140):\penalty0 1--67, 2020.

\bibitem[Rajpurkar(2016)]{squad}
Rajpurkar, P.
\newblock Squad: 100,000+ questions for machine comprehension of text.
\newblock \emph{arXiv preprint arXiv:1606.05250}, 2016.

\bibitem[Singh et~al.(2019)Singh, Natarajan, Shah, Jiang, Chen, Batra, Parikh,
  and Rohrbach]{singh2019towards}
Singh, A., Natarajan, V., Shah, M., Jiang, Y., Chen, X., Batra, D., Parikh, D.,
  and Rohrbach, M.
\newblock Towards vqa models that can read.
\newblock In \emph{Proceedings of the IEEE/CVF conference on computer vision
  and pattern recognition}, pp.\  8317--8326, 2019.

\bibitem[Sreenivas et~al.(2024)Sreenivas, Muralidharan, Joshi, Chochowski,
  Patwary, Shoeybi, Catanzaro, Kautz, and Molchanov]{nvidia_llm_distillation}
Sreenivas, S.~T., Muralidharan, S., Joshi, R., Chochowski, M., Patwary, M.,
  Shoeybi, M., Catanzaro, B., Kautz, J., and Molchanov, P.
\newblock Llm pruning and distillation in practice: The minitron approach.
\newblock \emph{arXiv preprint arXiv:2408.11796}, 2024.

\bibitem[Su et~al.(2024)Su, Ahmed, Lu, Pan, Bo, and Liu]{rope}
Su, J., Ahmed, M., Lu, Y., Pan, S., Bo, W., and Liu, Y.
\newblock Roformer: Enhanced transformer with rotary position embedding.
\newblock \emph{Neurocomputing}, 568:\penalty0 127063, 2024.

\bibitem[Sutton(2019)]{Sutton2019}
Sutton, R.~S.
\newblock The bitter lesson.
\newblock \url{http://www.incompleteideas.net/IncIdeas/BitterLesson.html},
  2019.
\newblock Accessed: 2023-11-07.

\bibitem[Tay et~al.(2021)Tay, Tran, Ruder, Gupta, Chung, Bahri, Qin,
  Baumgartner, Yu, and Metzler]{charformer}
Tay, Y., Tran, V.~Q., Ruder, S., Gupta, J., Chung, H.~W., Bahri, D., Qin, Z.,
  Baumgartner, S., Yu, C., and Metzler, D.
\newblock Charformer: Fast character transformers via gradient-based subword
  tokenization.
\newblock \emph{arXiv preprint arXiv:2106.12672}, 2021.

\bibitem[Tay et~al.(2022{\natexlab{a}})Tay, Dehghani, Tran, Garcia, Wei, Wang,
  Chung, Shakeri, Bahri, Schuster, et~al.]{ul2}
Tay, Y., Dehghani, M., Tran, V.~Q., Garcia, X., Wei, J., Wang, X., Chung,
  H.~W., Shakeri, S., Bahri, D., Schuster, T., et~al.
\newblock Ul2: Unifying language learning paradigms.
\newblock \emph{arXiv preprint arXiv:2205.05131}, 2022{\natexlab{a}}.

\bibitem[Tay et~al.(2022{\natexlab{b}})Tay, Wei, Chung, Tran, So, Shakeri,
  Garcia, Zheng, Rao, Chowdhery, et~al.]{upalm}
Tay, Y., Wei, J., Chung, H.~W., Tran, V.~Q., So, D.~R., Shakeri, S., Garcia,
  X., Zheng, H.~S., Rao, J., Chowdhery, A., et~al.
\newblock Transcending scaling laws with 0.1\% extra compute.
\newblock \emph{arXiv preprint arXiv:2210.11399}, 2022{\natexlab{b}}.

\bibitem[Team(2025)]{k15}
Team, K.
\newblock Kimi k1.5: Scaling reinforcement learning with llms.
\newblock \emph{Github}, 2025.

\bibitem[Touvron et~al.(2023)Touvron, Lavril, Izacard, Martinet, Lachaux,
  Lacroix, Rozi{\`e}re, Goyal, Hambro, Azhar, et~al.]{llama}
Touvron, H., Lavril, T., Izacard, G., Martinet, X., Lachaux, M.-A., Lacroix,
  T., Rozi{\`e}re, B., Goyal, N., Hambro, E., Azhar, F., et~al.
\newblock Llama: Open and efficient foundation language models.
\newblock \emph{arXiv preprint arXiv:2302.13971}, 2023.

\bibitem[Vaswani(2017)]{attention}
Vaswani, A.
\newblock Attention is all you need.
\newblock \emph{Advances in Neural Information Processing Systems}, 2017.

\bibitem[Wang et~al.(2022)Wang, Roberts, Hesslow, Le~Scao, Chung, Beltagy,
  Launay, and Raffel]{langarch}
Wang, T., Roberts, A., Hesslow, D., Le~Scao, T., Chung, H.~W., Beltagy, I.,
  Launay, J., and Raffel, C.
\newblock What language model architecture and pretraining objective works best
  for zero-shot generalization?
\newblock In \emph{International Conference on Machine Learning}, pp.\
  22964--22984. PMLR, 2022.

\bibitem[Warner et~al.(2024)Warner, Chaffin, Clavié, Weller, Hallström,
  Taghadouini, Gallagher, Biswas, Ladhak, Aarsen, Cooper, Adams, Howard, and
  Poli]{modern_bert}
Warner, B., Chaffin, A., Clavié, B., Weller, O., Hallström, O., Taghadouini,
  S., Gallagher, A., Biswas, R., Ladhak, F., Aarsen, T., Cooper, N., Adams, G.,
  Howard, J., and Poli, I.
\newblock Smarter, better, faster, longer: A modern bidirectional encoder for
  fast, memory efficient, and long context finetuning and inference, 2024.

\bibitem[Xiong et~al.(2020)Xiong, Yang, He, Zheng, Zheng, Xing, Zhang, Lan,
  Wang, and Liu]{prelayernorm}
Xiong, R., Yang, Y., He, D., Zheng, K., Zheng, S., Xing, C., Zhang, H., Lan,
  Y., Wang, L., and Liu, T.
\newblock On layer normalization in the transformer architecture.
\newblock In \emph{International Conference on Machine Learning}, pp.\
  10524--10533. PMLR, 2020.

\bibitem[Ye et~al.(2023)Ye, Hu, Xu, Ye, Yan, Xu, Li, Tian, Qian, Zhang,
  et~al.]{ye2023ureader}
Ye, J., Hu, A., Xu, H., Ye, Q., Yan, M., Xu, G., Li, C., Tian, J., Qian, Q.,
  Zhang, J., et~al.
\newblock Ureader: Universal ocr-free visually-situated language understanding
  with multimodal large language model.
\newblock \emph{arXiv preprint arXiv:2310.05126}, 2023.

\end{thebibliography}
\bibliographystyle{icml2025}

\end{document}